\pdfoutput=1

\documentclass[11pt]{article}

\usepackage{acl}

\usepackage{times}
\usepackage{latexsym}

\usepackage[T1]{fontenc}

\usepackage[utf8]{inputenc}

\usepackage{graphicx}
\usepackage{microtype}
\usepackage{amsmath}
\usepackage{amsthm}
\usepackage{booktabs}
\usepackage{algorithm}
\usepackage{algorithmic}
\usepackage{subfigure}
\usepackage{bm}
\usepackage{xspace}
\usepackage{multirow}
\usepackage{amssymb}
\usepackage[normalem]{ulem}
\usepackage{footmisc}
\usepackage{listings}
\usepackage{paralist}
\usepackage{enumitem}
\usepackage{epstopdf}
\usepackage{pifont}
\usepackage{xcolor}
\usepackage{mathrsfs}
\usepackage{textcomp}
\useunder{\uline}{\ul}{}
\usepackage{footmisc}
\usepackage{setspace}
\usepackage{marvosym}

\makeatletter
\def\@fnsymbol#1{}
\makeatother

\newcommand{\ie}{\emph{i.e.}, }

\newcommand{\eg}{\emph{e.g.}, }

\title{Context-Tuning: Learning Contextualized Prompts \\ for Natural Language Generation}

\author{
	Tianyi Tang\textsuperscript{\rm{1,4}},
	Junyi Li\textsuperscript{\rm{1,3}},
	Wayne Xin Zhao\textsuperscript{\rm{1,4,5  \Letter}\thanks{\textsuperscript{\Letter}\ Corresponding author}\ } \and
	Ji-Rong Wen\textsuperscript{\rm{1,2,4}} \\
	\textsuperscript{1}Gaoling School of Artificial Intelligence, Renmin University of China \\
	\textsuperscript{2}School of Information, Renmin University of China \\
	\textsuperscript{3}DIRO, Université de Montréal \\
	\textsuperscript{4}Beijing Key Laboratory of Big Data Management and Analysis Methods \\
	\textsuperscript{5}Beijing Academy of Artificial Intelligence, Beijing, 100084, China\\
	\texttt{steventianyitang@outlook.com lijunyi@ruc.edu.cn batmanfly@gmail.com} \\ 
}

\begin{document}
\maketitle
\begin{abstract}
Recently, pretrained language models (PLMs) have had exceptional success in language generation.
To leverage the rich knowledge encoded by PLMs, a simple yet powerful paradigm is to use \emph{prompts} in the form of either discrete tokens or continuous embeddings. In existing studies, these prompting methods are typically independent of the inputs, lacking sufficient consideration of input semantics.
To address this issue, we propose a novel continuous prompting approach, called \emph{context-tuning}, to fine-tuning PLMs for natural language generation. 
Firstly, the prompts are derived based on the input text to elicit useful knowledge from PLMs for generation. We refer to such prompts as \emph{contextualized prompts}.
Secondly, we use \emph{continuous inverse prompting} to improve the process of natural language generation by modeling an inverse generation process from output to input, making the generated text more relevant to the inputs. 
Furthermore, we utilize a lightweight context-tuning method that fine-tunes only 0.12\% of the parameters while maintaining good performance.
Our code is publicly available at \url{https://github.com/RUCAIBox/Context-Tuning}.
\end{abstract}

\section{Introduction}

Natural language generation (\emph{a.k.a.} text generation) aims to produce plausible and readable text in human language from input data~\cite{tg_survey}. 
Recently, large-scale pretrained language models (PLMs) such as BART~\cite{bart} have had exceptional success in language generation. To leverage the encoded knowledge from PLMs, prompting methods have been proposed~\cite{prompt-survey}, where the original input to PLMs has been extended by prepending discrete tokens or continuous embeddings (called \emph{prompts}). Following this paradigm, this work aims to study how to develop more effective prompting methods for text generation based on PLMs. 

\begin{table}[t]
\small
	\centering
	\resizebox{1\columnwidth}{!}{
		\begin{tabular}{lp{0.8\columnwidth}}
			\toprule
			\textbf{Title} & \uwave{Live-action medium} is inferior to \underline{animation medium} \\
			\cmidrule{1-2}
			\multicolumn{2}{l}{\textbf{Static Prompts}: \textit{Write a story about:} \textbf{Title}} \\
			\cmidrule{1-2}
			\multicolumn{2}{l}{\textbf{Contextualized Prompts}: $\bm{p}_{1 \cdots k}$ \ \textbf{Title} \  $\bm{p}_{k+1 \cdots 2k}$}  \\
			\midrule
			\textbf{Story} & I think that live-action works can't be considered art. They feel more like \uwave{documentaries} or \uwave{theater pieces} with \uwave{CGI} combined. The superiority of animated works is that they are more \underline{abstract} and \underline{imaginative} and characters show more \underline{emotion} and variety of \underline{designs}. \\
			\bottomrule                                            
	\end{tabular}}
	\caption{Example inputs (titles) and outputs (stories) of generation dataset \textsc{CMV}. Static prompts are human-written instructions, which are independent of input titles. $\bm{p}_{1 \cdots k}$ and $\bm{p}_{k+1 \cdots 2k}$ are contextualized prompts, which are derived conditioned on the input title. The wavy line and underline denote the corresponding information between input and output.}
	\label{tab:example}
\end{table}

Early methods focused on human-written (discrete) prompts by manually constructing task-specific prompt templates~\cite{t5,gpt2}, such as ``\texttt{TL;DR:}'' for the summarization task. 
Recent work has further proposed utilizing continuous prompts~\cite{prefix-tuning,prompttuning} for text generation. Continuous prompts consist of trainable parameters that do not correspond to real tokens and can be easily optimized during fine-tuning. 
However, existing prompting approaches typically adopt \emph{static prompts} for generation, \ie the prompts contain task-related information but remain the same for different input texts.

In this work, we mainly focus on challenging open-ended generation, such as story generation~\citep{fusion} and review generation~\citep{li-etal-2019-generating}. Under this setting, the input text usually contains very limited information, while the task goal is to generate an output sequence containing informative contents based on the given limited input.
The example in Table~\ref{tab:example} aims to generate a story about the topics of ``\emph{live-action}'' and ``\emph{animation}''. In such a case, it requires in-depth background knowledge about the two topics. As we can see, 
static prompts such as ``\textit{Write a story about:}'' are independent of the input title, making it difficult to capture the related aspects for this generation task.
Instead of static prompts, we argue that \emph{contextualized prompts} (as shown in Table~\ref{tab:example}) derived based on the input title will be more suited for this setting. 


To address the above issues, we propose \textbf{Context-Tuning}, a novel continuous prompting approach to fine-tuning PLMs for natural language generation. 
Our approach has three major technical contributions. 
Firstly, the prompts are derived based on input text to enrich the input by eliciting related knowledge from PLMs. Specifically, by concatenating limited input and a sequence of ``\texttt{[MASK]}'' tokens into BERT~\citep{bert}, we leverage its excellent mask-filling ability to predict these tokens, and the last hidden state of tokens can be used as prompt vectors. Since the prompts are highly related to the input context, we refer to them as \emph{contextualized prompts}. 
Secondly, to further enhance the relevance between the generated text and the input text, we extend inverse prompting~\cite{inverse} by incorporating continuous prompts. We refer to them as \emph{continuous inverse prompting}. By maximizing the likelihood of predicting inputs conditioned on the generated text and continuous prompts, context-tuning can generate texts highly relevant to the input text.
Moreover, to ease the training burden, we propose to use a lightweight context-tuning method~\citep{bitfit} that only fine-tunes the bias term of all model parameters. In this way, we can achieve a comparable performance (98.0\% of the full-tuned performance) by only tuning 0.12\% of the parameters, compared to full-tuned context-tuning.


To our knowledge, we are the first to encode input-related information into continuous prompts for text generation. Our context-tuning approach can elicit relevant knowledge according to specific input text and enhance the relevance between the generated text and the input text. We compare our method with several baseline models for the evaluation of four natural language generation tasks. Extensive experiments demonstrate the effectiveness of our proposed context-tuning approach.
\section{Related Work}

\begin{figure*}[t]
	\centering
	\includegraphics[width=0.95\textwidth]{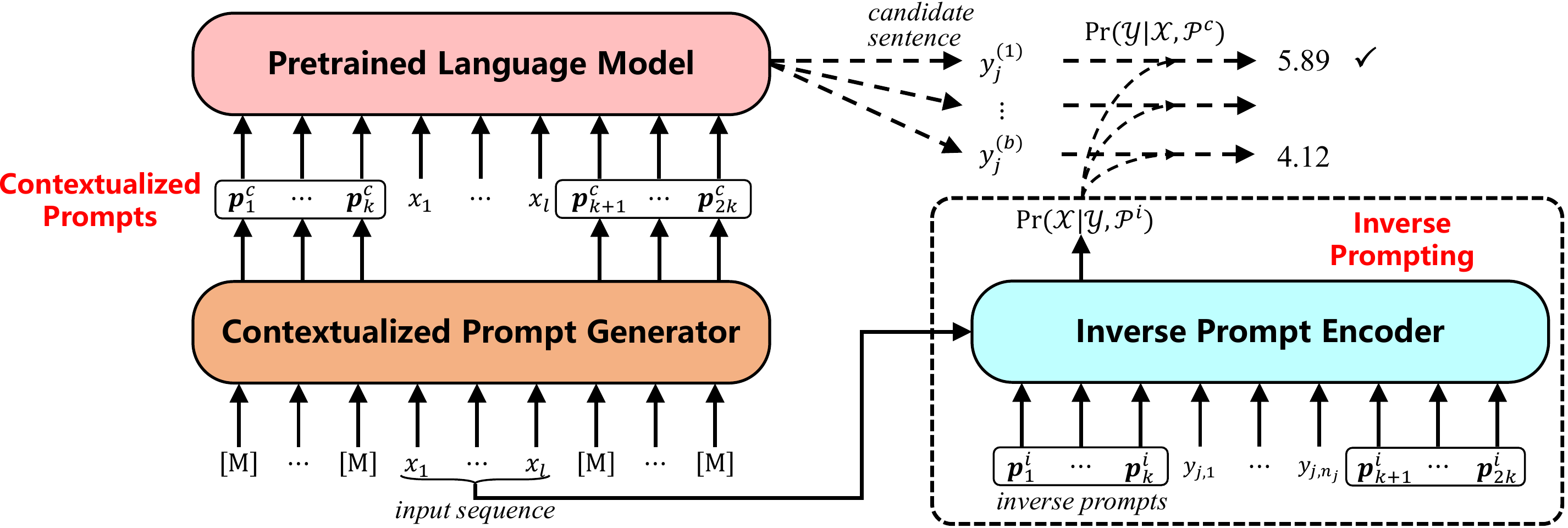}
	\caption{The overview of the proposed context-tuning. ``[M]'' denotes the mask token ``[MASK]''. By combining the forward probability $\text{Pr}(\mathcal{Y}|\mathcal{X},\mathcal{P}^c)$ (the left part) and backward probability $\text{Pr}(\mathcal{X}|\mathcal{Y},\mathcal{P}^i)$ (the right part), we select the sequence $y^{(i)}_j$ with the highest combined scores from all the candidates.}
	\label{fig:model}
\end{figure*}

\paragraph{Natural Language Generation.} 
Natural language generation is one of the most challenging fields in natural language processing (NLP). It aims to produce human-readable text from input text. Current state-of-the-art results for many generation tasks are based on fine-tuning PLMs, such as text summarization~\cite{bart}, dialogue system~\cite{dialogpt}, and data-to-text generation~\cite{Ribeiro}. 
As mentioned in~\citet{prompt-survey}, controlled text generation is relevant to our input-dependent method. The goal of controlled text generation is to direct the generated texts into specific styles~\cite{HuYLSX17}, lengths~\cite{KikuchiNSTO16}, or keywords~\cite{gsum}. In contrast, our contextualized prompts elicit knowledge from PLMs to enrich the input rather than control the specific properties of generated text.

\paragraph{Prompting Learning.}
Prompting  methods prepend  task instructions to the input and generate the output from PLMs. Most typical methods utilize manually designed task-specific prompts to adapt to different generation tasks~\cite{gpt2,t5}. However, it is time-consuming and laborious to construct human-written prompts for various generation tasks.
As a result, recent research has concentrated on automating the search for discrete prompts~\cite{autoprompt,lmbff}. Nonetheless, searching for prompts over discrete space is challenging to optimize due to the non-differentiable issues and continuous nature of neural networks. 
To handle these problems, many studies propose optimizing continuous prompts~\cite{prompttuning,prefix-tuning}, which are more expressive and flexible for any task. Among these works, prefix-tuning~\cite{prefix-tuning} and prompt tuning~\cite{prompttuning} are two representatives focused on text generation and natural language understanding (NLU) tasks, respectively. Compared with these continuous approaches, our context-tuning encodes the context information of inputs into the contextualized prompts and adopts continuous inverse prompting to enhance relevance further.

Most existing prompting methods~\citep{pet,autoprompt,prompttuning} focus on NLU tasks, which are choice questions that can easily be converted into filling ``\texttt{[MASK]}'' tasks. However, text generation aims to generate a sequence of tokens, in contrast to a few options in limited space. For example, prefix-tuning~\cite{prefix-tuning} and GENPET~\cite{genpet} have employed prompting methods for text generation. However, they mainly focus on lightweight fine-tuning or few-shot learning and do not achieve great performance under full tuning settings. In contrast, our context-tuning can improve performance under full tuning settings, and the lightweight strategy tunes only 0.2\% of the parameters while retaining good performance.
\section{The Proposed Approach}
In this section, we present the proposed \emph{context-tuning} to fine-tune PLMs for natural language generation. We first introduce the contextualized prompts based on the input text for generating informative text. To further enhance the relevance of the generated text to the input, we utilize continuous inverse prompting to enforce the prediction of inputs given the generated text and continuous prompts.
Figure~\ref{fig:model} presents an overall illustration of the proposed context-tuning approach.

For natural language generation, we consider a general task setting, where the model generates the output sequence $\mathcal{Y}$ conditioned on the input sequence $\mathcal{X}=\langle x_1,\dots,x_l \rangle$. The output text is usually composed of multiple sentences: $\mathcal{Y}=\{y_j : \langle y_{j,1},\dots,y_{j,t},\dots,y_{j,n_j} \rangle \}_{j=1}^m$. In context-tuning, we introduce contextualized prompts $\mathcal{P}^c=\langle \bm{p}^c_1,\dots,\bm{p}^c_k \rangle$ into the input side. Thus, the prompt-based generation task can be formulated as:
\begin{equation} \small
	\text{Pr}(\mathcal{Y}|\mathcal{X},\mathcal{P}^c)=\text{Pr}(y_1,\dots,y_m|x_1,\dots,x_l,\mathcal{P}^c).
\end{equation}

\subsection{Contextualized Prompts} \label{sec-cp}
Instead of static prompts~\cite{prompttuning} (\emph{irrelevant to input}), we use \emph{contextualized prompts}, which are expected to provide additional information, such as world knowledge, commonsense and task information extracted from PLMs to enrich the limited input text.

\paragraph{Masked Prompt Learning.} Specifically, unlike prefix-tuning~\cite{prefix-tuning}, which prepends a sequence of static vectors to each layer of PLMs, we append a sequence of $k$ continuous vectors on both the left and right sides of the input sequence $\mathcal{X}$ ($2k$ vectors in total). Inspired by the masked language modeling task of BERT~\citep{bert}, we use BERT as the prompt generator to derive the contextualized prompt vectors. We first place a sequence of $k$ ``\texttt{[MASK]}'' tokens on both sides of the input $\mathcal{X}$ as:
\begin{equation} \small
\widetilde{\mathcal{X}} = \texttt{[MASK]}_{1,\dots,k},\mathcal{X},\texttt{[MASK]}_{k+1,\dots,2k}.
\end{equation}
By feeding $\widetilde{\mathcal{X}}$ as the input of the prompt generator, we can obtain the top-layer representations of these ``\texttt{[MASK]}'' tokens:
\begin{equation} \small 
	\underbrace{\tilde{\bm{p}}^c_1,\dots,\tilde{\bm{p}}^c_k}_{\text{prefix prompts}}/\underbrace{\tilde{\bm{p}}^c_{k+1},\dots,\tilde{\bm{p}}^c_{2k}}_{\text{suffix prompts}} = \text{Prompt-Generator}(\widetilde{\mathcal{X}}).
\end{equation}
After shown in Section~\ref{sec-sen}, we set $k$ to $150$ with the best performance. Compared with randomly-initialized prompts, our BERT-based prompt learning method can better learn the dependency between the prompts and input texts.

\paragraph{Aligning to Word Embeddings.} Since these prompt vectors are latent embeddings, we further align them to the semantic space of word embeddings by designing a two-step \emph{semantic mapping} operator. For the first step, BERT predicts the probability distribution over its vocabulary based on these top-layer representations:
\begin{equation} \small
	\text{Pr}(w|\widetilde{\mathcal{X}})=\text{softmax}(\bm{W}^V\tilde{\bm{p}}^c_k), 
\end{equation}
where $\bm{W}^V$ is a trainable matrix. For the second step, we multiply the probability distribution $\text{Pr}(w|\widetilde{\mathcal{X}})$ with the word embedding matrix $\bm{E}$ and obtain the final contextualized prompt vectors:
\begin{equation}\label{eq-9} \small
	\bm{p}^c_k=\bm{E}\cdot\text{Pr}(w|\widetilde{\mathcal{X}}).
\end{equation}
We consider these mapped vectors as \emph{contextualized prompts}. 
Intuitively, the above \emph{semantic mapping} can be considered as a weighted average of word embeddings according to their probabilities. 
Compared with existing continuous prompts, our contextualized prompt vectors can better correspond to real word embeddings in semantic space, as shown in Section~\ref{sec:case-study}. 

\paragraph{Applying the Prompts.} After obtaining the contextualized prompts, we combine these prompt vectors and the word embeddings of $\mathcal{X}$ as the input of PLMs for generating the output text $\mathcal{Y}$. Specifically, we utilize BART as the base PLM to generate text by minimizing the cross-entropy loss function:

\begin{small}
	\begin{align} \label{eq-c}
		\mathcal{L}_c&=-\log \text{Pr}(\mathcal{Y}|\mathcal{X},\mathcal{P}^c) \\
		&=-\log \text{Pr}(\mathcal{Y}|\bm{p}^c_{1 \cdots k}, \bm{x}_{1\cdots l},\bm{p}^c_{k+1 \cdots 2k}), \nonumber
	\end{align}
\end{small}
where $\bm{p}^c_{1 \cdots k}$ denotes $\underline{\bm{p}^c_1,\dots,\bm{p}^c_k}$, $\bm{x}_{1\cdots l}$ denotes $\underline{\bm{x}_1,\dots,\bm{x}_l}$, and  $\bm{p}^c_{k+1 \cdots 2k}$ denotes $\underline{\bm{p}^c_{k+1},\dots,\bm{p}^c_{2k}}$.
By leveraging the encoded knowledge from PLM, the contextualized prompts are helpful to generate informative output texts.

\begin{algorithm}[t]
\small
	\begin{algorithmic}[1]
		\REQUIRE Model parameters $\Theta^{(c)}$ and $\Theta^{(i)}$, beam size $b$ and maximum number of sentences $n_m$ 
		\STATE \textbf{Input:} An input sequence $\mathcal{X}$
		\STATE \textbf{Output:} A generated sequence $\mathcal{Y}$
		\STATE Initialize step $j=0$
		\WHILE {$j<n_m$}
		\STATE Derive contextualized prompts $\mathcal{P}^c$ based on $\mathcal{X}$
		\STATE Generate $b$ candidate sentences $y_j^{(1)},\dots,y_j^{(b)}$ according to Eq.~\ref{eq-c}
		\STATE Utilize continuous inverse prompts $\mathcal{P}^i$ to compute the likelihood of candidate sentences according to Eq.~\ref{eq-i}
		\STATE Choose the best sentence as $y_s$ based on Eq.~\ref{eq-gen}
		\STATE Terminate the loop if $y_s$ contains the end of sentence token
		\STATE Update $j=j+1$
		\ENDWHILE
		\STATE Concatenate $y_1,\dots,y_j$ as generated sequence $\mathcal{Y}$
		\RETURN $\mathcal{Y}$
	\end{algorithmic}
	\caption{The algorithm procedure for generation process of context-tuning.}
	\label{alg-gen}
\end{algorithm}

\subsection{Continuous Inverse Prompting}
Although contextualized prompts can improve the informativeness of output, it still suffers from the off-topic generation issue as 
the text length increases~\cite{inverse}. To deal with this issue, we propose \emph{continuous inverse prompting} to enhance the relevance in an inverse manner from output to input. Compared to the previous inverse prompting that depends on artificial construction~\cite{inverse}, our inverse prompting is based on continuous prompts, which can be flexibly optimized during fine-tuning.

\paragraph{Output-to-Input Relevance Enhancement.} To model the relevance of output $\mathcal{Y}$ to input $\mathcal{X}$, we hypothesize that the output text is highly relevant to the input text if we can recover the input based on the output. Nevertheless, in some text generation tasks, it is non-intuitive to generate the input text given the output text. Hence, we utilize prompts to mitigate this issue. We introduce continuous inverse prompts $\mathcal{P}^i$ and append them on both sides of the output $\mathcal{Y}$. Then, we utilize another PLM to measure the conditional probability $\text{Pr}(\mathcal{X}|\mathcal{Y}, \mathcal{P}^i)$. Considering the output text $\mathcal{Y}$ might be much longer than the input text $\mathcal{X}$, we further model the probability at the sentence level:

\begin{small}
	\begin{align} \label{eq-i} \small
		\mathcal{L}_i&=-\log \text{Pr}(\mathcal{X}|\mathcal{Y},\mathcal{P}^i)  \\
		&=-\sum_{j=1}^m\log \text{Pr}(\mathcal{X}|\bm{p}^i_{1 \cdots k}, \bm{y}_{j,1}, \dots, \bm{y}_{j,n_j}, \bm{p}^i_{k+1 \cdots 2k}),\nonumber
	\end{align}
\end{small}
where $\bm{p}^i_{1 \cdots k}$ denotes $\underline{\bm{p}^i_1,\dots,\bm{p}^i_k}$ and  $\bm{p}^i_{k+1 \cdots 2k}$ denotes $\underline{\bm{p}^i_{k+1},\dots,\bm{p}^i_{2k}}$.
Unlike contextualized prompts in Section~\ref{sec-cp}, we expect inverse prompts to reflect better the relationship between $\mathcal{Y}$ and $\mathcal{X}$, which is dependent on the task rather than the input. Thus, the inverse prompts are static and continuous in our approach.

\paragraph{Generation with Inverse Prompting.} With the two techniques mentioned above in the generation process, we utilize a modified beam search algorithm shown in Algorithm~\ref{alg-gen} to generate the sequence $\mathcal{Y}$ with the highest combined probability:
\begin{equation} \label{eq-gen} \small
	\mathcal{Y}=\mathop{\text{argmax}}_{\mathcal{Y}} \log \text{Pr}(\mathcal{Y}|\mathcal{X},\mathcal{P}^c) + \lambda \log \text{Pr}(\mathcal{X}|\mathcal{Y},\mathcal{P}^i),
\end{equation}
where $\lambda$ is a hyper-parameter to balance these two probabilities. We set $\lambda$ to $4.0$ with the best balance of performance.

In contrast to contextualized prompts that enrich the input information, continuous inverse prompting makes the generation process more controllable. Even for latter generated sentences, it can still enforce them to adhere to the input topic.

\subsection{Discussion and Learning}
In this part, we present the model discussion and optimization.

\paragraph{Discussion and Comparison.} We use contextualized prompts (Eq.~\ref{eq-c}) to elicit useful knowledge from PLMs for different inputs. As a comparison, previous continuous prompting methods~\cite{prefix-tuning,prompttuning} adopt static prompts, which are irrelevant to the input. 
Besides, we propose continuous inverse prompting (Eq.~\ref{eq-i}) to enforce the relevance of long output text by considering a generation process from output to input. Different from the original inverse prompting, our inverse prompting is based on continuous prompts, which can be optimized during fine-tuning.

Considering that our method involves another PLM and more parameters, we propose a lightweight context-tuning approach. Following \citet{bitfit}, we only fine-tune the bias term of each parameter, resulting in fine-tuning only 0.12\% of the parameters of complete models. In the meanwhile, prefix-tuning~\cite{prefix-tuning} and prompt tuning~\cite{prompttuning} freeze the PLM and only fine-tune the parameters of prompts. Prefix-tuning fine-tunes prompts in each layer and tunes 16.4\% of the BART parameters, while prompt tuning only fine-tunes the prompt concatenated to the input, resulting in fine-tuning 0.05\% of the BART parameters.

\paragraph{Optimization.}
We use the base version of BERT as our prompt generator. The number of prompt vectors $k$ is set to $150$. We utilize the base version of BART for text generation. The hyper-parameter $\lambda$ in Eq.~\ref{eq-gen} is set to $4.0$. 
There are two sets of trainable parameters in contextualized prompts and continuous inverse prompting, denoted by $\Theta^{(c)}$ and $\Theta^{(i)}$, respectively.
First, we optimize $\Theta^{(c)}$, including BERT and BART, according to Eq.~\ref{eq-c}. Meanwhile, we optimize $\Theta^{(i)}$ according to the inverse generation loss using Eq.~\ref{eq-i}.
During inference, we combine them and select sentences that are both informative and relevant to the input text based on Algorithm~\ref{alg-gen} and Eq.~\ref{eq-gen}.

\section{Experiment}
In this section, we first set up the experiments and then report the results and analysis.

\begin{table}[t]
	\centering
	\resizebox{1\columnwidth}{!}{
		\begin{tabular}{r|rrrrr}
			\toprule
			\textbf{Dataset}        & \multicolumn{1}{c}{\textbf{\#Train}} & \multicolumn{1}{c}{\textbf{\#Valid}} & \multicolumn{1}{c}{\textbf{\#Test}} & \multicolumn{1}{c}{\textbf{\#Input}} & \multicolumn{1}{c}{\textbf{\#Output}} \\ \midrule
			\textsc{WP} & 53,516                               & 4,000                                & 2,000                               & 25.48                                & 150.64                                \\
			\textsc{ROC} & 176,688                               & 9,816                                & 4,909                               & 9.02                                & 40.72                                \\
			\textsc{CMV}      & 42,462                               & 6,480                                & 7,562                               & 17.89                               & 104.10                              \\
			\textsc{WikiP} & 69,288                               & 8,661                                & 8,662                               & 3.38                                & 194.72                                \\
			\bottomrule
	\end{tabular}}
	\caption{Statistics of our datasets after preprocessing. \#Train, \#Valid and, \#Test denote the number of examples in training, valid, and test datasets, respectively. \#Input and \#Output denote the average number of tokens in the input and output text.}
	\label{tab-data}
\end{table}

\begin{table*}[t]
	\centering
	
	\resizebox{1\textwidth}{!}{\begin{tabular}{r|rrrrrrrrc}
			\toprule
			\multicolumn{1}{c|}{\multirow{2.5}{*}{\textbf{Models}}} & \multicolumn{4}{c}{\textbf{\textsc{WritingPrompts}}} & \multicolumn{4}{c}{\textbf{\textsc{ROCStories}}} & \multicolumn{1}{c}{\multirow{2.5}{*}{\textbf{\#Para}}} \\ 
			\cmidrule(lr){2-5} \cmidrule(lr){6-9}
			& BLUE-$1$ & BLUE-$2$ & Dist-$1$ & Dist-$4$ & BLUE-$1$ & BLUE-$2$ & Dist-$1$ & Dist-$4$ &  \\ 
			\midrule
			& \multicolumn{9}{l}{\emph{Full fine-tuning}} \\
			\textbf{GPT-2}          &24.94& 9.03& 1.40&35.38&31.45&14.26& 2.21&58.63&1.2$\times$10\textsuperscript{8} \\
			\textbf{T5}             &20.76& 7.41& 1.25&27.77&31.31&14.23& 2.22&54.31&2.2$\times$10\textsuperscript{8} \\
			\textbf{BART}           &\underline{28.42}&\underline{11.31}& \underline{2.11}&\underline{62.05}&32.95&15.35& \underline{2.70}&68.88&1.4$\times$10\textsuperscript{8} \\
			\textbf{HINT}           &22.40& 8.40& --  &31.30&\underline{33.40}&\underline{15.40}& --  &\underline{69.30}&1.4$\times$10\textsuperscript{8} \\
			\textbf{Context-Tuning} &\textbf{29.88}&\textbf{11.85}& \textbf{2.49}&\textbf{67.78}&\textbf{34.65}&\textbf{16.60}& \textbf{3.16}&\textbf{75.53}&2.5$\times$10\textsuperscript{8} \\
			\midrule[0.3pt]
			& \multicolumn{9}{l}{\emph{Lightweight fine-tuning}} \\
			\textbf{Prompt Tuning}  &16.26& 5.18& \textbf{3.71}&\textbf{69.68}&27.27&10.49& 2.44&62.12&7.6$\times$10\textsuperscript{4} \\
			\textbf{Prefix-Tuning}  &\underline{28.39}&\underline{10.76}& 1.72&58.34&\underline{30.62}&\underline{13.51}& \textbf{2.51}&\underline{67.19}&2.3$\times$10\textsuperscript{7} \\
			\textbf{Context-Tuning} &\textbf{29.45}&\textbf{10.90}& \underline{1.78}&\underline{62.89}&\textbf{32.24}&\textbf{14.30}& \underline{2.49}&\textbf{68.92}&3.0$\times$10\textsuperscript{5} \\
			\midrule
			& \multicolumn{4}{c}{\textbf{\textsc{ChangeMyView}}} & \multicolumn{4}{c}{\textbf{\textsc{Wikiplots}}} & \\ 
			\cmidrule(lr){2-5} \cmidrule(lr){6-9}
			& BLUE-$1$ & BLUE-$2$ & Dist-$1$ & Dist-$4$ & BLUE-$1$ & BLUE-$2$ & Dist-$1$ & Dist-$4$ &  \\ 
			\midrule
			& \multicolumn{9}{l}{\emph{Full fine-tuning}} \\
			\textbf{GPT-2}          &23.39& 8.32& 0.75&37.18&23.74& 9.33& 0.90&38.39&1.2$\times$10\textsuperscript{8} \\
			\textbf{T5}             &20.89& 7.79& \underline{1.01}&42.48&14.83& 6.09& 1.33&39.25&2.2$\times$10\textsuperscript{8} \\
			\textbf{BART}           &\underline{25.69}& \underline{9.77}& \textbf{1.11}&\textbf{61.21}&27.12&11.54& \underline{1.82}&\underline{49.54}&1.4$\times$10\textsuperscript{8} \\
			\textbf{Context-Tuning} &\textbf{26.11}&\textbf{10.00}& 0.99&\underline{57.38}&\textbf{27.80}&\textbf{11.82}&\textbf{2.05}&\textbf{51.96}
			&2.5$\times$10\textsuperscript{8} \\
			\midrule[0.3pt]
			& \multicolumn{9}{l}{\emph{Lightweight fine-tuning}} \\
			\textbf{Prompt Tuning}  &22.54& 8.11& \textbf{1.43}&\textbf{64.60}&18.64& 6.98& 2.78&58.53&7.6$\times$10\textsuperscript{4} \\
			\textbf{Prefix-Tuning}  &\underline{25.72}& \underline{9.84}& 0.96&54.66&\textbf{27.30}&\textbf{11.60}& \underline{1.95}&\underline{51.05}&2.3$\times$10\textsuperscript{7} \\
			\textbf{Context-Tuning} &\textbf{28.83}&\textbf{10.96}& \underline{0.97}&\underline{57.79}&\underline{26.93}&\underline{11.53}& \textbf{2.48}&\textbf{59.34}&3.0$\times$10\textsuperscript{5} \\
			\bottomrule
	\end{tabular}}
	\caption{Performance comparison of different methods for open-ended text generation tasks. Dist-$n$ is short for Distinct-$n$. Bold and underlined fonts denote the best and second-best methods (the same below). \#Para denotes the number of fine-tuned parameters in each method. The results of HINT are from its original paper~\citep{hint}. ``--'' means HINT does not compute the corresponding result.}
	\label{tab:main-results}
\end{table*}

\subsection{Experimental Setup}

\subsubsection{Construction of the Datasets} To measure the performance of our proposed context-tuning, we evaluate it on four open-ended text generation tasks: \textsc{WritingPrompts} (\textsc{WP}), \textsc{ROCStories} (\textsc{ROC}), \textsc{ChangeMyView} (\textsc{CMV}), and \textsc{Wikiplots} (\textsc{WikiP}). 
Specifically, \textsc{WP}~\citep{fusion} consists of pairs of story premises and responses from the WritingPrompts forum. \textsc{ROC}~\citep{roc} is a dataset consisting of five-sentence commonsense stories. Here, we use the first sentence as the input to generate the following four sentences. For \textsc{WP} and \textsc{ROC}, we utilize the version provided by~\citet{hint} for a fair comparison.
\textsc{CMV}~\citep{cmv} contains pairs of post statements on a controversial issue, which are collected from Reddit. 
\textsc{WikiP}\footnote{\url{https://github.com/markriedl/WikiPlots}} is a collection of story plots from Wikipedia. We use the sub-header word to generate the full story.

Since some dataset outputs are significantly long, we discard examples where the text contains more than 512 tokens due to the length limitation of PLMs. We summarize the statistics of four datasets after preprocessing in Table~\ref{tab-data}.

\subsubsection{Baseline Methods} We consider the following baselines as comparisons: GPT-2, BART, T5, HINT, prefix-tuning, and prompt tuning. Among these baselines, GPT-2~\cite{gpt2}, BART~\cite{bart}, and T5~\cite{t5} are three prevalent PLMs for natural language generation; HINT~\citep{hint} is a strong baseline model specially designed for generating long and coherent texts; prefix-tuning~\cite{prefix-tuning} and prompt tuning~\cite{prompttuning} are the recently proposed lightweight models using continuous prompts for generation tasks. We utilize the base version for all PLMs for a fair comparison.

\subsubsection{Implementation Details}
For fine-tuning settings, we consider two strategies: full fine-tuning and lightweight fine-tuning, to compare our methods with different baselines. In the lightweight fine-tuning settings, our context-tuning only tunes the bias term of each parameter.

In all experiments, we utilize the Adam optimizer and set $\beta_1=0.9$, $\beta_2=0.999$, $\epsilon=1 \times 10^{-8}$. We train our model for $20$ epochs and utilize the model with the best performance on the validation set for generation. During inference, we apply the nucleus sampling with $p=0.9$ and temperature of $0.7$. We train our model using NVIDIA A100 GPUs on Ubuntu 18.04 and employ the NLP open-source library Transformers~\citep{huggingface} and text generation library TextBox~\citep{textbox}.

\subsubsection{Evaluation Metrics} To evaluate the performance of different methods of natural language generation, we adopt two automatic evaluation metrics, including BLEU~\cite{bleu} and Distinct~\cite{distinct}. Specifically, BLEU evaluates the quality of generated and real text, while Distinct measures the diversity of generated texts.

\subsection{Performance Comparison}
We present the results of different methods on generation tasks in Table~\ref{tab:main-results}.

First, we can see that BART performs best compared to other PLMs on these generation tasks. Pretrained on the large-scale corpus, PLMs can better understand natural language and fluently express human language. We consider the better performance of BART is due to the encoder-decoder architecture and the DAE pretraining task. That is the major reason we adopt BART as our base generation model.

Second, the recently proposed continuous prompting methods, prefix-tuning, and prompt tuning do not achieve ideal performance in these tasks. This finding shows  that natural language generation tasks are more challenging than NLU tasks. Only fine-tuning a few parameters cannot outperform full fine-tuning. 

Finally, our model outperforms all the baselines (including the strong baseline HINT) over four tasks under both full and lightweight tuning settings. The reason is that our context-tuning utilizes \emph{contextualized prompts}, which can serve as queries to elicit input-relevant knowledge from PLMs. Under the lightweight fine-tuning settings, our context-tuning has superior results to prefix-tuning, only with 1.3\% of the parameters of prefix-tuning and 0.2\% of the parameters of BART. Some of the performance under lightweight settings can even outperform the full tuning, which may be a solution to catastrophic forgetting~\citep{kadapter}. And a major reason is that prefix-tuning and prompt tuning adopt static prompts, which are task-specific and unrelated to the context information.

\begin{table}[t]
	\centering
	\resizebox{1\columnwidth}{!}{
		\begin{tabular}{l|rrrr}
			\toprule
			\textbf{Models}  & \multicolumn{1}{c}{\textbf{B-$1$}}  & \multicolumn{1}{c}{\textbf{B-$2$}} & \multicolumn{1}{c}{\textbf{D-$1$}} & \multicolumn{1}{c}{\textbf{D-$4$}} \\ 
			\midrule
			\textbf{Context-Tuning}                         &29.88&11.85& 2.49&67.78             \\
			\midrule
			w/o Continuous w \textbf{Manual}                &                                     &                                    &                                     \\
			$\quad$ - human-written prompt\textsubscript{1} &27.96&10.93&1.92&59.97 \\
			$\quad$ - human-written prompt\textsubscript{2} &29.14&11.56&2.09&61.69 \\ 
			w/o BERT w \textbf{RoBERTa}                     &27.31&10.73&1.56&57.51                               \\
			w/o \textbf{Semantic Mapping}                   &29.19&11.33&1.72&58.57   \\
			w/o \textbf{Inverse Prompting}                  &29.31&11.45&2.19&64.09 \\
			\bottomrule
	\end{tabular}}
	\caption{Ablation analysis on \textsc{WritingPrompts} dataset.}
	\label{tab:ablation-results}
\end{table}

\subsection{Ablation Analysis}
In this part, we construct ablation experiments to test the effectiveness of our proposed context-tuning. 
In contrast to previous prompt-based studies, our context-tuning has made several improvements. First, compared with manual prompts, we propose a continuous prompting approach to fine-tuning PLMs. Second, we adopt BERT as the prompt generator to derive the contextualized prompt vectors with semantic mapping. Finally, we utilize inverse prompting to enhance the relevance of the generated texts further. Here, we would like to examine how each factor contributes to the final performance. To see this, we prepare several variants for a comparison: 

$\bullet$ \emph{w/o Continuous w Manual}: the variant removes the continuous prompts but utilizes two kinds of human-written prompts, \ie prompt\textsubscript{1}: ``\texttt{Title: \textbf{\$Input} Story:}'' and prompt\textsubscript{2}: ``\texttt{Given the title \textbf{\$Input}, please write the following story:}''. 

$\bullet$ \emph{w/o BERT w RoBERTa}: the variant replaces BERT with RoBERTa~\cite{roberta} to form the prompt generator.

$\bullet$ \emph{w/o Semantic Mapping}: the variant does not align to word embeddings and directly utilizes the top-layer representations of ``\texttt{[MASK]}'' tokens in the prompt generator.

$\bullet$ \emph{w/o Inverse Prompting}: the variant removes inverse prompting (Eq.~\ref{eq-gen}) from our proposed context-tuning.

\begin{table}[t]
	\centering
	\resizebox{1\columnwidth}{!}{
		\begin{tabular}{r|r|rrrr}
			\toprule
			\textbf{Models}  & \multicolumn{1}{c|}{\textbf{TT (\%)}} & \multicolumn{1}{c}{\textbf{Flu.}} & \multicolumn{1}{c}{\textbf{Info.}} & \multicolumn{1}{c}{\textbf{Rel.}} & \multicolumn{1}{c}{\textbf{Coh.}} \\ \midrule
			\textbf{GPT-2}    & {\ul 81.20}                     & {\ul 3.90}                           & {\ul 3.27}                                   & {\ul 3.77}                            & 3.50                                   \\
			\textbf{T5}      & 61.48                           & 3.58                                 & 3.02                                         & 3.64                                  & 3.25                                   \\
			\textbf{BART}    & 77.17                           & 3.82                                 & {\ul 3.27}                                   & 3.74                                  & {\ul 3.59}                             \\
			\textbf{Context-Tuning} & \textbf{82.83}                  & \textbf{4.12}                        & \textbf{3.47}                                & \textbf{3.94}                         & \textbf{3.85}                          \\
			\cmidrule(lr){2-6}
			\textbf{Gold}    & 94.00                           & 4.26                                 & 3.90                                         & 4.33                                  & 4.01                                   \\ 
			\bottomrule
	\end{tabular}}
	\caption{Turing test (TT) and human evaluation on \textsc{WritingPrompts}. ``Gold'' indicates the ground-truth texts. Flu., Info., Rel., and Coh. denote fluency, informativeness, relevance, and coherence, respectively.}
	\label{tab:human-results}
\end{table}

From Table~\ref{tab:ablation-results}, we can see that variants replacing continuous prompts with manual prompts are worse than the model with continuous prompts. The performance of manual prompts is sensitive to different instructions and does not always lead to gains. This verifies the effectiveness of utilizing continuous prompts rather than discrete ones for text generation tasks.
The variants replacing the BERT-based prompt generator with RoBERTa are worse than the full model. 
We further observe a slight performance drop when our method removes the semantic mapping and inverse prompting. This implies that the proposed semantic mapping and continuous inverse prompting approaches can enforce the informativeness relevance of output text. 

\begin{table}[t]
	\centering
		\begin{tabular}{l|rrrr}
			\toprule
			\textbf{\#Prompt}  & \multicolumn{1}{c}{\textbf{B-$1$}}  & \multicolumn{1}{c}{\textbf{B-$2$}} & \multicolumn{1}{c}{\textbf{D-$1$}} & \multicolumn{1}{c}{\textbf{D-$4$}} \\ 
			\midrule
			$k=50$ &27.82&11.00&2.28&63.77 \\
			$k=100$&28.55&11.12&2.17&63.69 \\
			$k=150$&29.31&11.45&2.19&64.09 \\
			$k=200$&28.48&11.22&1.97&61.09 \\
			\bottomrule
	\end{tabular}
	\caption{Performance tuning on \textsc{WritingPrompts} dataset. We do not utilize continuous inverse prompting methods here.}
	\label{tab:parameter-tuning}
\end{table}

\subsection{Model Sensitivity}  \label{sec-sen}
In this part, we construct sensitivity analyses \emph{w.r.t.} the number $k$ of prompt vectors on the \textsc{WritingPrompts} dataset.

In contextualized prompt learning, the number of prompt vectors is a key factor that influences the performance of our model. A longer sequence of prompt vectors means more trainable parameters and, therefore, more expressive power. Here, we will examine how it affects the final performance of our context-tuning. Given the statistics in Table~\ref{tab-data}, we vary the number of prompt vectors in the set $\{50, 100, 150, 200\}$. We separately train our model with different numbers of prompt vectors and do not utilize continuous inverse prompting methods for convenience. As shown in Table~\ref{tab:parameter-tuning}, the performance of our model gradually improves as the number of prompt vectors increases up to a threshold, and then a performance drop occurs. More importantly, our model achieves the best performance with 150 prompt vectors over baselines.

\begin{table*}[t]
	\scriptsize
	\centering
		\begin{tabular}{p{0.05\textwidth}p{0.9\textwidth}}
			\toprule
			\textbf{Title} & Write a Nature Documentary about Pokémon . \\
			\midrule
			\textbf{GPT-2} & $<$newline$>$ it's been a long time since i've seen one of these things. $<$newline$>$ i don't know how long it has been, but i know that it will be long. i have seen it before, and i will see it again. it is the only thing i can think of that makes sense to me, because it makes me feel like i am part of something greater than myself. and that is why i want to know more about it. because if i could, i would know what it was like to be human. to have a sense of purpose and purposelessness, to feel that there is a purpose in everything that you do, even if you are just a shell of what you once were. but that does not make it any less of a feeling. that feeling that i feel when i look into the eyes of another human, when they look at me and say, `` i love you. '' \\
			\cmidrule{1-2}
			\textbf{BART} & the pokémon are the most powerful of all theokémon in the world. $<$newline$>$ they are the best of the best, but they are also the most dangerous. they have the ability to move faster than any otherokémon on the planet, and they are able to communicate with each other in a way that is impossible for humans to understand. however, they are not able to fight each other, because they are unable to communicate in a language that is not their own. they can only communicate through their own language, which is very different from any other language that humans can understand. they are very difficult to understand, and can not communicate in any language that can be understood by humans. they also have a very limited ability to speak, so they can not understand the language that they are speaking. in fact, they have no ability to communicate at all, because their language is very similar to that of a human. \\
			\midrule
			\midrule
			\multicolumn{2}{l}{\textbf{Closer words to contextualized prompts}: \uline{forest} \uline{trees} \uline{marching} \uline{winds} \uline{food} \dotuline{historically} \dotuline{story} \dotuline{movie} \uwave{pet} \uwave{pikachu} \uwave{animation} \dashuline{whilst} \dashuline{critical} \dashuline{asking} \dashuline{write}} \\
			\midrule
			\textbf{Context-Tuning} & `` \uwave{pokémon}! '' $<$newline$>$ `` what are you talking about? '' \dotuline{the man asked.} `` i'm talking about a \uwave{pokéball}. it's called \uwave{pikachu}, and it lives in the \uline{forest}. it lives on the edge of the \uline{forest}, where it can \uline{eat} anything it wants to eat. it is the only thing that can keep it alive. it can live in the \uline{woods}, but it can also be eaten by any other \uline{creature}. it has the ability to eat anything that it wants, and can even eat any other \uline{animal} it wants. '' \dotuline{he said}. the man looked at the man, and said, `` i don't know what you're talking about, but i do know that it can be eaten. '' the two men looked at each other, and \dotuline{the man spoke}, `` you've got to be kidding me. '' \\
			\bottomrule
	\end{tabular}
	\caption{The generated examples of the given title from the \textsc{WritingPrompts} dataset. Marks in closer words to contextualized prompts and texts generated by context-tuning refer to four chosen aspects, \ie nature, documentary, Pokémon, and the story generation task.}
	\label{tab:examples}
\end{table*}

\subsection{Human Evaluation} \label{sec:human}
Besides  automatic evaluation,  we further conduct a human evaluation for testing the effectiveness of our approach.
We randomly select 500 input texts from the test set of the \textsc{WritingPrompts} dataset. We collect the stories generated by GPT-2, BART, T5, and context-tuning, then shuffle them for human evaluation. Following~\citet{inverse}, we invite ten human judges to assign scores to a generated text concerning four factors of quality, namely 
\emph{informativeness} (how much it provides valuable and meaningful information), 
\emph{relevance} (how relevant it is according to the input contexts), \emph{coherence} evaluates (how coherent both intra and inter sentences are) and \emph{fluency} (how likely a human produces the generated text).

We adopt a 5-point Likert scale as the scoring mechanism, in which 5-point means ``very satisfying'', and 1-point means ``very terrible''. Furthermore, inspired by~\citet{inverse}, we design a Turing test where a human judge is asked to distinguish whether a human produces the given text. The detailed evaluation guidelines and examples are listed in Figure~\ref{fig:guide}, Figure~\ref{fig:fluency}, Figure~\ref{fig:info}, and Figure~\ref{fig:coher} in the Appendix.

We present the human evaluation results in Table \ref{tab:human-results}. It can be seen that our model is better than the three baselines with a large margin. The major reason is that we utilize the contextualized prompts derived from the input text. Our contextualized prompts can extract knowledge from PLMs and serve as additional input information to be fed into PLMs, which improves the informativeness of the generated text. Moreover, the proposed continuous inverse prompting method enhances the relevance of the generated text to the input.

\subsection{Qualitative Analysis}\label{sec:case-study}
In this part, we present an intuitive analysis of why our model works well.

Table~\ref{tab:examples} presents an example story from the \textsc{WritingPrompts} dataset and the generated story by our model and two baselines, \ie GPT-2 and BART. As we can see, there is limited information in the input premise, besides several keywords such as nature, documentary, and Pokémon. 

First, we can see that the story generated by our context-tuning is highly relevant to the input text and conveys richer semantic information. A primary reason might be that our contextualized prompts can elicit input-relevant knowledge from PLMs for generating more informative text. 
Although PLMs perform well in generating fluent text, we can see that GPT-2 and BART are still prone to generating unmeaningful and irrelevant content, such as ``I love you'' and ``language''. 

Furthermore, to probe whether our contextualized prompts contain input-relevant knowledge, we find close actual words for a better explanation. 
We use $2k$ contextualized prompts in total, and for each continuous prompt, we recall the word in the BERT vocabulary with the closest cosine distance to it. Finally, we select some words from $2k$ recalled words and showcase them grouped by four aspects in the row \emph{closer words to contextualized prompts} of Table~\ref{tab:examples}.
As we can see, most recalled keywords are included in the story generated by our context-tuning. It shows that our contextualized prompts can better capture input-relevant knowledge. For example, the keywords ``forest'', ``woods'', and ``animal'' are closely related to the aspect of \emph{nature}.

\section{Conclusion}
This paper presents a novel continuous prompting approach, \ie context-tuning, to fine-tuning PLMs for natural language generation. The core idea is to inject input-related context information into continuous prompts, called contextualized prompts, to enhance the informativeness of generation. The contextualized prompts can elicit input-relevant knowledge from PLMs to enrich the input text. Furthermore, to enhance the relevance of the generated text to the inputs, we adopt a continuous inverse prompting method to refine the forward generation process by modeling an inverse generation process from output to input. We also propose a lightweight method for efficient training. Extensive experiments on four generation tasks have demonstrated the effectiveness of our model in fine-tuning PLMs for text generation tasks. 

In future work, we will consider integrating more types of context information (\eg sentiment) to derive more expressive prompts and investigate how our model could be applied to other tasks.

\section*{Acknowledgement}
This work was partially supported by Beijing Natural Science Foundation under Grant No. 4222027,  Beijing Outstanding Young Scientist Program under Grant No. BJJWZYJH012019100020098 and Beijing Academy of Artificial Intelligence (BAAI). Xin Zhao is the corresponding author.

\bibliography{reference}

\begin{thebibliography}{30}
\expandafter\ifx\csname natexlab\endcsname\relax\def\natexlab#1{#1}\fi

\bibitem[{Ben~Zaken et~al.(2022)Ben~Zaken, Goldberg, and Ravfogel}]{bitfit}
Elad Ben~Zaken, Yoav Goldberg, and Shauli Ravfogel. 2022.
\newblock \href {https://aclanthology.org/2022.acl-short.1} {{B}it{F}it: Simple
  parameter-efficient fine-tuning for transformer-based masked
  language-models}.
\newblock In \emph{Proceedings of the 60th Annual Meeting of the Association
  for Computational Linguistics (Volume 2: Short Papers)}, pages 1--9, Dublin,
  Ireland. Association for Computational Linguistics.

\bibitem[{Devlin et~al.(2019)Devlin, Chang, Lee, and Toutanova}]{bert}
Jacob Devlin, Ming-Wei Chang, Kenton Lee, and Kristina Toutanova. 2019.
\newblock \href {https://doi.org/10.18653/v1/N19-1423} {{BERT}: Pre-training of
  deep bidirectional transformers for language understanding}.
\newblock In \emph{Proceedings of the 2019 Conference of the North {A}merican
  Chapter of the Association for Computational Linguistics: Human Language
  Technologies, Volume 1 (Long and Short Papers)}, pages 4171--4186,
  Minneapolis, Minnesota. Association for Computational Linguistics.

\bibitem[{Dou et~al.(2021)Dou, Liu, Hayashi, Jiang, and Neubig}]{gsum}
Zi-Yi Dou, Pengfei Liu, Hiroaki Hayashi, Zhengbao Jiang, and Graham Neubig.
  2021.
\newblock \href {https://doi.org/10.18653/v1/2021.naacl-main.384} {{GS}um: A
  general framework for guided neural abstractive summarization}.
\newblock In \emph{Proceedings of the 2021 Conference of the North American
  Chapter of the Association for Computational Linguistics: Human Language
  Technologies}, pages 4830--4842, Online. Association for Computational
  Linguistics.

\bibitem[{Fan et~al.(2018)Fan, Lewis, and Dauphin}]{fusion}
Angela Fan, Mike Lewis, and Yann Dauphin. 2018.
\newblock \href {https://doi.org/10.18653/v1/P18-1082} {Hierarchical neural
  story generation}.
\newblock In \emph{Proceedings of the 56th Annual Meeting of the Association
  for Computational Linguistics (Volume 1: Long Papers)}, pages 889--898,
  Melbourne, Australia. Association for Computational Linguistics.

\bibitem[{Gao et~al.(2021)Gao, Fisch, and Chen}]{lmbff}
Tianyu Gao, Adam Fisch, and Danqi Chen. 2021.
\newblock \href {https://doi.org/10.18653/v1/2021.acl-long.295} {Making
  pre-trained language models better few-shot learners}.
\newblock In \emph{Proceedings of the 59th Annual Meeting of the Association
  for Computational Linguistics and the 11th International Joint Conference on
  Natural Language Processing (Volume 1: Long Papers)}, pages 3816--3830,
  Online. Association for Computational Linguistics.

\bibitem[{Guan et~al.(2021)Guan, Mao, Fan, Liu, Ding, and Huang}]{hint}
Jian Guan, Xiaoxi Mao, Changjie Fan, Zitao Liu, Wenbiao Ding, and Minlie Huang.
  2021.
\newblock \href {https://doi.org/10.18653/v1/2021.acl-long.499} {Long text
  generation by modeling sentence-level and discourse-level coherence}.
\newblock In \emph{Proceedings of the 59th Annual Meeting of the Association
  for Computational Linguistics and the 11th International Joint Conference on
  Natural Language Processing (Volume 1: Long Papers)}, pages 6379--6393,
  Online. Association for Computational Linguistics.

\bibitem[{Hu et~al.(2017)Hu, Yang, Liang, Salakhutdinov, and Xing}]{HuYLSX17}
Zhiting Hu, Zichao Yang, Xiaodan Liang, Ruslan Salakhutdinov, and Eric~P. Xing.
  2017.
\newblock \href {https://proceedings.mlr.press/v70/hu17e.html} {Toward
  controlled generation of text}.
\newblock In \emph{Proceedings of the 34th International Conference on Machine
  Learning}, volume~70 of \emph{Proceedings of Machine Learning Research},
  pages 1587--1596. PMLR.

\bibitem[{Hua and Wang(2020)}]{cmv}
Xinyu Hua and Lu~Wang. 2020.
\newblock \href {https://doi.org/10.18653/v1/2020.emnlp-main.57} {{PAIR}:
  Planning and iterative refinement in pre-trained transformers for long text
  generation}.
\newblock In \emph{Proceedings of the 2020 Conference on Empirical Methods in
  Natural Language Processing (EMNLP)}, pages 781--793, Online. Association for
  Computational Linguistics.

\bibitem[{Kikuchi et~al.(2016)Kikuchi, Neubig, Sasano, Takamura, and
  Okumura}]{KikuchiNSTO16}
Yuta Kikuchi, Graham Neubig, Ryohei Sasano, Hiroya Takamura, and Manabu
  Okumura. 2016.
\newblock \href {https://doi.org/10.18653/v1/D16-1140} {Controlling output
  length in neural encoder-decoders}.
\newblock In \emph{Proceedings of the 2016 Conference on Empirical Methods in
  Natural Language Processing}, pages 1328--1338, Austin, Texas. Association
  for Computational Linguistics.

\bibitem[{Lester et~al.(2021)Lester, Al-Rfou, and Constant}]{prompttuning}
Brian Lester, Rami Al-Rfou, and Noah Constant. 2021.
\newblock \href {https://doi.org/10.18653/v1/2021.emnlp-main.243} {The power of
  scale for parameter-efficient prompt tuning}.
\newblock In \emph{Proceedings of the 2021 Conference on Empirical Methods in
  Natural Language Processing}, pages 3045--3059, Online and Punta Cana,
  Dominican Republic. Association for Computational Linguistics.

\bibitem[{Lewis et~al.(2020)Lewis, Liu, Goyal, Ghazvininejad, Mohamed, Levy,
  Stoyanov, and Zettlemoyer}]{bart}
Mike Lewis, Yinhan Liu, Naman Goyal, Marjan Ghazvininejad, Abdelrahman Mohamed,
  Omer Levy, Veselin Stoyanov, and Luke Zettlemoyer. 2020.
\newblock \href {https://doi.org/10.18653/v1/2020.acl-main.703} {{BART}:
  Denoising sequence-to-sequence pre-training for natural language generation,
  translation, and comprehension}.
\newblock In \emph{Proceedings of the 58th Annual Meeting of the Association
  for Computational Linguistics}, pages 7871--7880, Online. Association for
  Computational Linguistics.

\bibitem[{Li et~al.(2016)Li, Galley, Brockett, Gao, and Dolan}]{distinct}
Jiwei Li, Michel Galley, Chris Brockett, Jianfeng Gao, and Bill Dolan. 2016.
\newblock \href {https://doi.org/10.18653/v1/N16-1014} {A diversity-promoting
  objective function for neural conversation models}.
\newblock In \emph{Proceedings of the 2016 Conference of the North {A}merican
  Chapter of the Association for Computational Linguistics: Human Language
  Technologies}, pages 110--119, San Diego, California. Association for
  Computational Linguistics.

\bibitem[{Li et~al.(2021)Li, Tang, He, Jiang, Hu, Xie, Chen, Yu, Zhao, and
  Wen}]{textbox}
Junyi Li, Tianyi Tang, Gaole He, Jinhao Jiang, Xiaoxuan Hu, Puzhao Xie, Zhipeng
  Chen, Zhuohao Yu, Wayne~Xin Zhao, and Ji-Rong Wen. 2021.
\newblock \href {https://doi.org/10.18653/v1/2021.acl-demo.4} {{T}ext{B}ox: A
  unified, modularized, and extensible framework for text generation}.
\newblock In \emph{Proceedings of the 59th Annual Meeting of the Association
  for Computational Linguistics and the 11th International Joint Conference on
  Natural Language Processing: System Demonstrations}, pages 30--39, Online.
  Association for Computational Linguistics.

\bibitem[{Li et~al.(2022)Li, Tang, Zhao, Nie, and Wen}]{tg_survey}
Junyi Li, Tianyi Tang, Wayne~Xin Zhao, Jian-Yun Nie, and Ji-Rong Wen. 2022.
\newblock \href {https://arxiv.org/abs/2201.05273} {A survey of pretrained
  language models based text generation}.
\newblock \emph{arXiv preprint arXiv:2201.05273}.

\bibitem[{Li et~al.(2019)Li, Zhao, Wen, and Song}]{li-etal-2019-generating}
Junyi Li, Wayne~Xin Zhao, Ji-Rong Wen, and Yang Song. 2019.
\newblock \href {https://doi.org/10.18653/v1/P19-1190} {Generating long and
  informative reviews with aspect-aware coarse-to-fine decoding}.
\newblock In \emph{Proceedings of the 57th Annual Meeting of the Association
  for Computational Linguistics}, pages 1969--1979, Florence, Italy.
  Association for Computational Linguistics.

\bibitem[{Li and Liang(2021)}]{prefix-tuning}
Xiang~Lisa Li and Percy Liang. 2021.
\newblock \href {https://doi.org/10.18653/v1/2021.acl-long.353} {Prefix-tuning:
  Optimizing continuous prompts for generation}.
\newblock In \emph{Proceedings of the 59th Annual Meeting of the Association
  for Computational Linguistics and the 11th International Joint Conference on
  Natural Language Processing (Volume 1: Long Papers)}, pages 4582--4597,
  Online. Association for Computational Linguistics.

\bibitem[{Liu et~al.(2021)Liu, Yuan, Fu, Jiang, Hayashi, and
  Neubig}]{prompt-survey}
Pengfei Liu, Weizhe Yuan, Jinlan Fu, Zhengbao Jiang, Hiroaki Hayashi, and
  Graham Neubig. 2021.
\newblock \href {http://arxiv.org/abs/2107.13586} {Pre-train, prompt, and
  predict: A systematic survey of prompting methods in natural language
  processing}.
\newblock \emph{arXiv preprint arXiv:2107.13586}.

\bibitem[{Liu et~al.(2019)Liu, Ott, Goyal, Du, Joshi, Chen, Levy, Lewis,
  Zettlemoyer, and Stoyanov}]{roberta}
Yinhan Liu, Myle Ott, Naman Goyal, Jingfei Du, Mandar Joshi, Danqi Chen, Omer
  Levy, Mike Lewis, Luke Zettlemoyer, and Veselin Stoyanov. 2019.
\newblock \href {http://arxiv.org/abs/1910.10683} {Roberta: A robustly
  optimized bert pretraining approach}.
\newblock \emph{arXiv preprint arXiv:1907.11692}.

\bibitem[{Mostafazadeh et~al.(2016)Mostafazadeh, Chambers, He, Parikh, Batra,
  Vanderwende, Kohli, and Allen}]{roc}
Nasrin Mostafazadeh, Nathanael Chambers, Xiaodong He, Devi Parikh, Dhruv Batra,
  Lucy Vanderwende, Pushmeet Kohli, and James Allen. 2016.
\newblock \href {https://doi.org/10.18653/v1/N16-1098} {A corpus and cloze
  evaluation for deeper understanding of commonsense stories}.
\newblock In \emph{Proceedings of the 2016 Conference of the North {A}merican
  Chapter of the Association for Computational Linguistics: Human Language
  Technologies}, pages 839--849, San Diego, California. Association for
  Computational Linguistics.

\bibitem[{Papineni et~al.(2002)Papineni, Roukos, Ward, and Zhu}]{bleu}
Kishore Papineni, Salim Roukos, Todd Ward, and Wei-Jing Zhu. 2002.
\newblock \href {https://doi.org/10.3115/1073083.1073135} {{B}leu: a method for
  automatic evaluation of machine translation}.
\newblock In \emph{Proceedings of the 40th Annual Meeting of the Association
  for Computational Linguistics}, pages 311--318, Philadelphia, Pennsylvania,
  USA. Association for Computational Linguistics.

\bibitem[{Radford et~al.(2019)Radford, Wu, Child, Luan, Amodei, Sutskever
  et~al.}]{gpt2}
Alec Radford, Jeffrey Wu, Rewon Child, David Luan, Dario Amodei, Ilya
  Sutskever, et~al. 2019.
\newblock \href
  {https://cdn.openai.com/better-language-models/language_models_are_unsupervised_multitask_learners.pdf}
  {Language models are unsupervised multitask learners}.
\newblock \emph{OpenAI blog}, 1(8):9.

\bibitem[{Raffel et~al.(2020)Raffel, Shazeer, Roberts, Lee, Narang, Matena,
  Zhou, Li, and Liu}]{t5}
Colin Raffel, Noam Shazeer, Adam Roberts, Katherine Lee, Sharan Narang, Michael
  Matena, Yanqi Zhou, Wei Li, and Peter~J. Liu. 2020.
\newblock \href {http://jmlr.org/papers/v21/20-074.html} {Exploring the limits
  of transfer learning with a unified text-to-text transformer}.
\newblock \emph{Journal of Machine Learning Research}, 21(140):1--67.

\bibitem[{Ribeiro et~al.(2021)Ribeiro, Schmitt, Sch{\"u}tze, and
  Gurevych}]{Ribeiro}
Leonardo F.~R. Ribeiro, Martin Schmitt, Hinrich Sch{\"u}tze, and Iryna
  Gurevych. 2021.
\newblock \href {https://doi.org/10.18653/v1/2021.nlp4convai-1.20}
  {Investigating pretrained language models for graph-to-text generation}.
\newblock In \emph{Proceedings of the 3rd Workshop on Natural Language
  Processing for Conversational AI}, pages 211--227, Online. Association for
  Computational Linguistics.

\bibitem[{Schick and Sch{\"u}tze(2021{\natexlab{a}})}]{pet}
Timo Schick and Hinrich Sch{\"u}tze. 2021{\natexlab{a}}.
\newblock \href {https://doi.org/10.18653/v1/2021.eacl-main.20} {Exploiting
  cloze-questions for few-shot text classification and natural language
  inference}.
\newblock In \emph{Proceedings of the 16th Conference of the European Chapter
  of the Association for Computational Linguistics: Main Volume}, pages
  255--269, Online. Association for Computational Linguistics.

\bibitem[{Schick and Sch{\"u}tze(2021{\natexlab{b}})}]{genpet}
Timo Schick and Hinrich Sch{\"u}tze. 2021{\natexlab{b}}.
\newblock \href {https://doi.org/10.18653/v1/2021.emnlp-main.32} {Few-shot text
  generation with natural language instructions}.
\newblock In \emph{Proceedings of the 2021 Conference on Empirical Methods in
  Natural Language Processing}, pages 390--402, Online and Punta Cana,
  Dominican Republic. Association for Computational Linguistics.

\bibitem[{Shin et~al.(2020)Shin, Razeghi, Logan~IV, Wallace, and
  Singh}]{autoprompt}
Taylor Shin, Yasaman Razeghi, Robert~L. Logan~IV, Eric Wallace, and Sameer
  Singh. 2020.
\newblock \href {https://doi.org/10.18653/v1/2020.emnlp-main.346}
  {{A}uto{P}rompt: {E}liciting {K}nowledge from {L}anguage {M}odels with
  {A}utomatically {G}enerated {P}rompts}.
\newblock In \emph{Proceedings of the 2020 Conference on Empirical Methods in
  Natural Language Processing (EMNLP)}, pages 4222--4235, Online. Association
  for Computational Linguistics.

\bibitem[{Wang et~al.(2021)Wang, Tang, Duan, Wei, Huang, Ji, Cao, Jiang, and
  Zhou}]{kadapter}
Ruize Wang, Duyu Tang, Nan Duan, Zhongyu Wei, Xuanjing Huang, Jianshu Ji,
  Guihong Cao, Daxin Jiang, and Ming Zhou. 2021.
\newblock \href {https://doi.org/10.18653/v1/2021.findings-acl.121}
  {{K-Adapter}: {I}nfusing {K}nowledge into {P}re-{T}rained {M}odels with
  {A}dapters}.
\newblock In \emph{Findings of the Association for Computational Linguistics:
  ACL-IJCNLP 2021}, pages 1405--1418, Online. Association for Computational
  Linguistics.

\bibitem[{Wolf et~al.(2020)Wolf, Debut, Sanh, Chaumond, Delangue, Moi, Cistac,
  Rault, Louf, Funtowicz, Davison, Shleifer, von Platen, Ma, Jernite, Plu, Xu,
  Le~Scao, Gugger, Drame, Lhoest, and Rush}]{huggingface}
Thomas Wolf, Lysandre Debut, Victor Sanh, Julien Chaumond, Clement Delangue,
  Anthony Moi, Pierric Cistac, Tim Rault, Remi Louf, Morgan Funtowicz, Joe
  Davison, Sam Shleifer, Patrick von Platen, Clara Ma, Yacine Jernite, Julien
  Plu, Canwen Xu, Teven Le~Scao, Sylvain Gugger, Mariama Drame, Quentin Lhoest,
  and Alexander Rush. 2020.
\newblock \href {https://doi.org/10.18653/v1/2020.emnlp-demos.6} {Transformers:
  State-of-the-art natural language processing}.
\newblock In \emph{Proceedings of the 2020 Conference on Empirical Methods in
  Natural Language Processing: System Demonstrations}, pages 38--45, Online.
  Association for Computational Linguistics.

\bibitem[{Zhang et~al.(2020)Zhang, Sun, Galley, Chen, Brockett, Gao, Gao, Liu,
  and Dolan}]{dialogpt}
Yizhe Zhang, Siqi Sun, Michel Galley, Yen-Chun Chen, Chris Brockett, Xiang Gao,
  Jianfeng Gao, Jingjing Liu, and Bill Dolan. 2020.
\newblock \href {https://doi.org/10.18653/v1/2020.acl-demos.30} {{DIALOGPT} :
  Large-scale generative pre-training for conversational response generation}.
\newblock In \emph{Proceedings of the 58th Annual Meeting of the Association
  for Computational Linguistics: System Demonstrations}, pages 270--278,
  Online. Association for Computational Linguistics.

\bibitem[{Zou et~al.(2021)Zou, Yin, Zhong, Yang, Yang, and Tang}]{inverse}
Xu~Zou, Da~Yin, Qingyang Zhong, Hongxia Yang, Zhilin Yang, and Jie Tang. 2021.
\newblock \href {https://doi.org/10.1145/3447548.3467418} {Controllable
  generation from pre-trained language models via inverse prompting}.
\newblock In \emph{Proceedings of the 27th ACM SIGKDD Conference on Knowledge
  Discovery \& Data Mining}, KDD '21, page 2450–2460. Association for
  Computing Machinery.

\end{thebibliography}

\newpage

\appendix

\begin{figure*}[t]
	\centering
	\begin{tabular}{|p{135mm}|}
		\hline
		\small
		
		Thank you for taking time out of your busy schedule to participate in our scientific research evaluation!
		
		Our research work is to let the machine generate corresponding story, for a given title, and hope that it is as close as possible to what humans write. Hence, we need to evaluate whether it meets the standards that people think.
		
		In this task, you will see a title, idea or introduction such as:
		
		\textit{You have a very shitty type of precognition .}
		
		Then you'll see a corresponding story or comment, either written by a human or a machine, but you don't know which one was written, and you'll have to rate it based on the following standard:
		
		\begin{itemize}
			\item \textbf{Turing Test:} whether the text was written by a human;
			\item \textbf{Fluency:} whether the text has good form, logical reading and smooth sentences;
			\item \textbf{Informative:} whether the text contains meaningful content and will not be boring to read;
			\item \textbf{Relevance:} whether the text is highly relevant to the input context;
			\item \textbf{Coherence:} whether the logic is coherent and not contradictory;
		\end{itemize}
		
		For Turing Test, just judge the text by instinct. For the other four factors, we adopt a 5-point likert scale as the scoring mechanism, in which 5-point means ``very satisfying'', and 1-point means ``very terrible''.
		
		\\
		\hline
	\end{tabular}
	\caption{Evaluation guidelines for \textsc{WritingPrompts} dataset.}
	\label{fig:guide}
\end{figure*}

\begin{figure*}[t]
	\centering
	\begin{tabular}{|p{135mm}|}
		\hline
		\small
		\textbf{Fluency:} whether the text has good form, logical reading and smooth sentences.
		
		\begin{itemize}
			\item \textbf{1 point example:} I was a exterminator on earth. {\ul I killed every insect and animal I killed in my life.} If I win, I go to heaven, lose, and go to hell. I was an exterminator in hell. I was the exterminator of the world. {\ul I exterminated every insect, animal, and animal that I killed.} {\ul I didn't have to kill them all.} I just had to do it. {\ul I had to kill all of them.} I couldn't let them get away with it. {\ul So I killed them all, and I did it all over again. And again and again.} I don't know how long I've been doing this, but it's been a long time. \textbf{(containing some grammatical errors and repetitive contents)}
			\item \textbf{1 point example:} I followed you into the dark . We carved our names into the bark .  `` {\ul Forever and always Through short nights and through long days .} '' {\ul But that tree withered so long Its voice died in sweet silent song .} Dead and gone We left and lost our life bond . I always remember your words Of comfort and joy like the birds . Sweet and sad ; It was all I never had . For I carved both names on that tree No kind woman would be with me . No more hope ; You and me , the end of a rope You were not ever real and I know that now . `` Always to love you '' , I end with that vow . \textbf{(there is no continuity between the words of the sentence, and the content is intermittent)}
			\item \textbf{3 point example:} I've been trying to kill my master for years. I've tried to kill him for years, but he's always been there for me. He's the only one who knows what I'm going to do, and I don't care. {\ul I 'll kill him if I have to. But I can't do it anymore.} {\ul I haven't been able to do it for years now. I can not do it any more.} I just want to go back to my master. I want to be with him again. But he won't let me go back. I know it's not fair, but I just need to get back to him. \textbf{(each sentence is grammatically correct and fluent, but contains certain repetitions and discontinuities in semantics)}
			\item \textbf{3 point example:} It's been a long time since I've seen her. She's always been there for me. I'm not sure how long I have been here, but I know she's here. I know I 'll never see her again. I don't know if she 'll ever see me again. But I know it's time. I can feel it in my bones, in my skin, in the bones of my bones. I can't help but think of her. I remember her when I first met her, when I was young. She was so beautiful, so full of life. I couldn't wait to meet her again, to see her smile again. \textbf{(sentences are fluent, but similar words are used repeatedly in the sentence, resulting in ambiguous meaning and confusing)}
			\item \textbf{5 point example:} Long ago his heart had warmed , three thousand years - long enough to mourn , the deeds of past and of damnation , stripped of humanity and of his station . He resided in the pits of hell the oldest friend of satan , waiting as the centuries pass watching hells inflation , resting on brimstone as passing devils chatter and laugh , who is this old man and what sin has made him . a curious young man with a glint in his eye asks his sentence , and with creaks and groans the old man rose for the first time in ages , he look at the spirit and with a heavy sigh he came out with , I 'm god and I made this . 
			\item \textbf{5 point example:} Tell us your faults ? Really ? This was the question - the shibboleth - that unlocked the cosmos ? The Masters could have picked a scientist to answer but they feared she might mask ignorance . They could have picked from our global leaders bit they feared that they would mask deceit . They could have picked a holy man but feared he would mask violence , oppression , hate , intolerance ... the list of disqualifying sins was almost too long to enumerate . So they picked Josh Thornton , a 45 year old MBA in human resources . `` Our greatest weakness ? Well , I think we work a little too hard and , as a race , we might be a bit of a perfectionist . ''
		\end{itemize}
		\\
		\hline
	\end{tabular}
	\caption{Evaluation examples for \emph{fluency} factor.}
	\label{fig:fluency}
\end{figure*}

\begin{figure*}[t]
	\centering
	\begin{tabular}{|p{135mm}|}
		\hline
		\small
		\textbf{Informative:} whether the text contains meaningful content and will not be boring to read.
		
		\begin{itemize}
			\item \textbf{1 point example:} I've been doing this for years. I've been doing this for years. I've been doing this for years. I've been doing this for years. I've been doing this for years. I've been doing this for years. I've been doing this for years. I've been doing this for years. I've been doing this for years. I've been doing this for years. I've been doing this for years. I've been doing this for years. I've been doing this for years. I've been doing this for years. I've been doing this for years. \textbf{(sentence repetition)}
			\item \textbf{1 point example:} `` I'm the First ****bender. ''  `` What's that? '' I asked. `` I don't know, '' he replied. `` But you're the first ****bender, right? '' `` Yeah, '' I said. `` You're not the first. '' `` Well, I guess that's true. '' He paused. `` What do you mean? '' He looked at me. `` That's right. '' I looked at him. `` It's not like you've ever been the first... '' He trailed off. `` No, it's just... '' I trailed off, `` I've never been the First... '' \textbf{(conversation sentences are very short, and they are all safe replies)}
			\item \textbf{3 point example:} `` What the hell is this? ''  `` I don't know, '' I said. `` I'm not sure what to do with this. I've been doing this for a long time, and it's not like I 'll be able to do anything about it. '' He looked at me with a confused look on his face. `` What the fuck? '' I asked. `` You're not going to do this again. '' I replied. `` It's just a matter of time, '' he said. I looked at the frying pan again. `` Don't worry, it 'll work. It 'll do. '' `` What's the problem? '' He asked. I nodded. `` Well, I guess I 'd better get out of here. '' The frying pan buzzed at me and text appeared reading `` level 18 cooking required to use object. '' \textbf{(containing rich dialogue, but the content of the dialogue is meaningless)}
			\item \textbf{3 point example:} The stars fell like rain, and we danced. The moon was bright, and the stars danced. The stars were bright, but the stars were not bright. They were bright and the sky was blue. We danced, but we didn't know what to do. We danced and we sang and we laughed and we talked and we cried and we screamed and we played and we giggled and we thought we were going to die, but it wasn't. It was just the stars falling like rain. And we danced, and it was beautiful. \textbf{(the article has a certain content at the beginning, but it is more verbose at the end)}
			\item \textbf{5 point example:} She is the woman you have loved since the day you met her . Everyday she has a smile on her face beautiful as ever . You love her but are afraid of what she will say when you tell her . It was raining and you ran for shelter , a small roof at the bus stop . Tired and panting you barely notice her sitting beside you , she calls your name . You jump a little bit and become nervous when you recognize her . You stare at each other not knowing what to do or say , and then she kisses you . An alarm sounds , you wake up in your room all alone , another dream \textbf{(a whole story)}
			\item \textbf{5 point example:} I feel like it 's worth pointing out that a lot of these are kind of situational . I think all of these rules are good to follow if what you 're writing is something you 're trying to submit to someone formally or something , but a lot of them are not important in casual writing ( such as someone 's dialogue or something like that ) . {\ul For example , `` literally '' has been used for hyperbole for a very long time} -- I know I 've heard that Nathaniel Hawthorne did it , and I hardly think he was the first . It pisses a lot of people off but it 's not like it 's a new phenomenon and it 's not like it 's a corruption of the language . Things about spelling and homonyms and stuff should probably be followed just about all the time though . \textbf{(using concrete examples)}
		\end{itemize}
		\\
		\hline
	\end{tabular}
	\caption{Evaluation examples for \emph{informative} factor.}
	\label{fig:info}
\end{figure*}

\begin{figure*}[t]
	\centering
	\begin{tabular}{|p{135mm}|}
		\hline
		\small
		\textbf{Coherence} evaluates how content is coherent considering both intra- and inter-sentence correlation.
		
		\begin{itemize}
			\item \textbf{1 point example:} {\ul i don't know if it's a good thing or a bad thing, but i've found that if i'm going to write a story, i need to be able to get it out of my head.} i think that's the most important thing. if i want to write something, i have to know what i want it to be. i have a lot of ideas, but they're just not good enough. i'll try to find a way to get them out, but if i can't find the right words to write, i will probably have to go back and re - write it. \textbf{(no semantic connection between sentences)}
			\item \textbf{1 point example:} `` I'm sorry, '' I said to myself. `` I don't know what to do. '' I replied. `` I just want to see you again. '' I looked at my reflection, and it was the same. I couldn't tell if it was a reflection or a reflection, but I knew that it was my reflection. I looked back at the mirror, and I saw that I was the one who was going to be my future soulmate. I smiled, and said, `` I love you, '' and walked away.
			\item \textbf{3 point example:} I have a very shitty type of precognition. I don't know what it is, but it's something I have to do. I'm not sure why I do it, but I do. I've been doing it for years now, and I haven't been able to figure out why. It's not like I have any control over it. {\ul I can't control what I do}, or what I say, or {\ul how I act}. {\ul I can only control the way I act}, how I react, and how I feel. {\ul I have no control over my actions, and no control of my emotions.} {\ul I just have to control my emotions,} and that's all I can do. \textbf{(after careful consideration, many logical contradictions were found)}
			\item \textbf{3 point example:} I 'd like to take a moment to appreciate Sir Terry Pratchett. I've read a lot of his work, and I'm not sure if it's because of his writing style or because of the way he wrote it. {\ul I don't know if he's a good writer, or if he is a bad writer}, but I do know that he is one of the best writers I have ever read. I think that's why I love him so much. {\ul I also think that he has a great sense of humor, and that he doesn't have a bad sense of humour.} \textbf{(some repeated information, but other content is ok)}
			\item \textbf{5 point example:} You eagerly await your pizza to come because you ordered from this new Italian {\ul Pizza} owed by two brother , you remember that one of their names are Mario but you forgot the other . The {\ul Pizza finally} arrives a bit late from this tall guy dressed in green . You pay him take , take the pizza but forget to tip . When you start eating you get a bit dizzy so you lay down and fall asleep quite quickly . You wake up in a in a place covered in mushrooms with a little man dressed as a mushroom telling you that `` You need to save the princess '' . \textbf{(smooth connection between context)}
			\item \textbf{5 point example:} When 1st purge happened , no one thought people would attack each other . A desperate party know only as Al Queda broke the rules and decided that it would do what no one else would have done . Bomb Manhattan . That single move destroyed not only the Republicans and the Democrats , it also destroyed morale . Hundreds of fully armed fat Politicians fled to the streets , screaming out jibberish and shooting anyone they see . Millions lay dead as all parties Jump onto their jets towards Manhattan , preparing to be included in the giant Cesspit of a war know as the Purge . When the Morning came . There were no victors . Only that the red dawn came and claimed .
		\end{itemize}
		\\
		\hline
	\end{tabular}
	\caption{Evaluation examples for \emph{coherence} factor.}
	\label{fig:coher}
\end{figure*}

\end{document}